\documentclass[10pt, a4paper]{article}

\usepackage[hyphens,spaces]{url}
\usepackage[usenames,dvipsnames,table]{xcolor}
\usepackage{booktabs}
\usepackage{enumitem}

\usepackage{lrec}
\usepackage{graphicx}
\usepackage{tabularx}
\usepackage{soul}
\usepackage{epstopdf}
\usepackage[utf8]{inputenc}
\usepackage{hyperref}
\usepackage{xstring}
\usepackage{multirow}
\usepackage{bigstrut}
\usepackage{multicol}
\usepackage{subcaption}
\usepackage{float}
\usepackage{collcell}

\usepackage{pgfplots}
\pgfplotsset{compat=1.14}   
\usepgfplotslibrary{colorbrewer}
\usetikzlibrary{pgfplots.groupplots}
\pgfplotsset{
    cycle list/Dark2,
    cycle multiindex* list={
        mark list*\nextlist
        Dark2\nextlist
    },
}

\newcommand{\ApplyGradient}[1]{
    \ifdim #1 pt > 0.75 pt
        \pgfmathsetmacro{\PercentColor}{100.0*#1}
        \hspace{-0.33em}\colorbox{RoyalPurple!\PercentColor!white}{\textcolor{white}{#1}}
    \else
        \pgfmathsetmacro{\PercentColor}{100.0*#1}
        \hspace{-0.33em}\colorbox{RoyalPurple!\PercentColor!white}{\textcolor{black}{#1}}
    \fi
}

\newcolumntype{G}{>{\collectcell\ApplyGradient}c<{\endcollectcell}}

\usepackage{adjustbox}

\newcolumntype{R}[2]{%
    >{\adjustbox{angle=#1,lap=\width-(#2)}\bgroup}%
    l%
    <{\egroup}%
}
\newcommand*\rotz{\multicolumn{1}{R{45}{-0.33em}}}

\newcommand{\fc}[1]{\textbf{#1}\ \ }

\newcommand{\snubes}{\textsc{NUBes-PHI}}

\title{Sensitive Data Detection and Classification in Spanish Clinical Text: Experiments with BERT}

\name{Aitor Garc\'ia-Pablos, Naiara Perez, Montse Cuadros}

\address{SNLT group at Vicomtech Foundation, Basque Research and Technology Alliance (BRTA) \\
         Donostia/San-Sebastián, 20009, Spain \\
         \{agarciap, nperez, mcuadros\}@vicomtech.org\\}

\abstract{Massive digital data processing provides a wide range of opportunities and benefits, but at the cost of endangering personal data privacy. Anonymisation consists in removing or replacing sensitive information from data, enabling its exploitation for different purposes while preserving the privacy of individuals. Over the years, a lot of automatic anonymisation systems have been proposed; however, depending on the type of data, the target language or the availability of training documents, the task remains challenging still. The emergence of novel deep-learning models during the last two years has brought large improvements to the state of the art in the field of Natural Language Processing. These advancements have been most noticeably led by BERT, a model proposed by Google in 2018, and the shared language models pre-trained on millions of documents.
In this paper, we use a BERT-based sequence labelling model to conduct a series of anonymisation experiments on several clinical datasets in Spanish. We also compare BERT to other algorithms. The experiments show that a simple BERT-based model with general-domain pre-training obtains highly competitive results without any domain specific feature engineering. \\
\newline }

\begin{document}

\maketitleabstract

%
%
\section{Introduction}
\label{sec:introduction}

During the first two decades of the 21st century, the sharing and processing of vast amounts of data has become pervasive. This expansion of data sharing and processing capabilities is both a blessing and a curse. Data helps build better information systems for the digital era and enables further research for advanced data management that benefits the society in general. But the use of this very data containing sensitive information conflicts with private data protection, both from an ethical and a legal perspective.

There are several application domains on which this situation is particularly acute. This is the case of the medical domain \cite{Abouelmehdi2018}. There are plenty of potential applications for advanced medical data management that can only be researched and developed using real data; yet, the use of medical data is severely limited --when not entirely prohibited-- due to data privacy protection policies.

One way of circumventing this problem is to anonymise the data by removing, replacing or obfuscating the personal information mentioned, as exemplified in Table \ref{tab:anon-examples}. This task can be done by hand, having people read and anonymise the documents one by one. Despite being a reliable and simple solution, this approach is tedious, expensive, time consuming and difficult to scale to the potentially thousands or millions of documents that need to be anonymised.

\begin{table*}[!hbtp]
    \centering
    \begin{tabular}{rl}
        \toprule
        original & \small \texttt{Paciente de \textbf{64 años} operado de una hernia el \textbf{12/01/2016} por \textbf{la Dra Lopez}} \\
        \midrule
        example 1 & \small  \texttt{Paciente de \textbf{XXXXXXX} operado de una hernia el \textbf{XXXXXXXXXX} por \textbf{XXXXXXXXXXXX}} \\
        example 2 & \small  \texttt{Paciente de \textbf{[-AGE-]} operado de una hernia el \textbf{[--DATE--]} por \textbf{[--DOCTOR--]}} \\
        example 3 & \small  \texttt{Paciente de \textbf{59 años} operado de una hernia el \textbf{05/06/2019} por \textbf{el Dr Sancho}} \\ 
        \bottomrule
    \end{tabular}
    \caption{Anonymization examples of ``64-year-old patient operated on a hernia on the 12/01/2016 by Dr Lopez''; sensitive data and their substitutions are highlighted in bold.}
    \label{tab:anon-examples}
\end{table*}

For this reason, numerous of systems and approaches have been developed during the last decades to attempt to automate the anonymisation of sensitive content, starting with the automatic detection and classification of sensitive information. Some of these systems rely on rules, patterns and dictionaries, while others use more advanced techniques related to machine learning and, more recently, deep learning.

Given that this paper is concerned with text documents (e.g. medical records), the involved techniques are related to Natural Language Processing (NLP). When using NLP approaches, it is common to pose the problem of document anonymisation as a sequence labelling problem, i.e. classifying each token within a sequence as being sensitive information or not. Further, depending on the objective of the anonymisation task, it is also important to determine the type of sensitive information (names of individuals, addresses, age, sex, etc.).

The anonymisation systems based on NLP techniques perform reasonably well, but are far from perfect. Depending on the difficulty posed by each dataset or the amount of available data for training machine learning models, the performance achieved by these methods is not enough to fully rely on them in certain situations \cite{Abouelmehdi2018}. However, in the last two years, the NLP community has reached an important milestone thanks to the appearance of the so-called Transformers neural network architectures \cite{wolf2019transformers}. 
In this paper, we conduct several experiments in sensitive information detection and classification on Spanish clinical text using BERT (from `Bidirectional Encoder Representations from Transformers') \cite{devlin2018bert} as the base for a sequence labelling approach. The experiments are carried out on two datasets: the MEDDOCAN: \textit{Medical Document Anonymization} shared task dataset \cite{marimon2019automatic}, and \textsc{NUBes} \cite{lima2019a}, a corpus of real medical reports in Spanish. In these experiments, we compare the performance of BERT with other machine-learning-based systems,  some of which use language-specific features. Our aim is to evaluate how good a BERT-based model performs without language nor domain specialisation apart from the training data labelled for the task at hand.

The rest of the paper is structured as follows: the next section describes related work about data anonymisation in general and clinical data anonymisation in particular; it also provides a more detailed explanation and background about the Transformers architecture and BERT. Section \ref{sec:materials-and-methods} describes the data involved in the experiments and the systems evaluated in this paper, including the BERT-based system; finally, it details the experimental design. Section \ref{sec:results} introduces the results for each set of experiments. Finally, Section \ref{sec:conclusions} contains the conclusions and future lines of work.

%
%
\section{Related Work}
\label{sec:related}

The state of the art in the field of Natural Language Processing (NLP) has reached an important milestone in the last couple of years thanks to deep-learning architectures, increasing in several points the performance of new models for almost any text processing task.

The major change started with the Transformers model proposed by \newcite{vaswani2017attention}. It substituted the widely used recurrent and convolutional neural network architectures by another approach based solely on self-attention, obtaining an impressive performance gain. The original proposal was focused on an encoder-decoder architecture for machine translation, but soon the use of Transformers was made more general \cite{wolf2019transformers}.
There are several other popular models that use Transformers, such as Open AI's GPT and GPT2 \cite{radford2019language}, RoBERTa \cite{liu2019roberta} and the most recent XLNet \cite{yang2019xlnet}; still, BERT \cite{devlin2018bert} is one of the most widespread Transformer-based models.

BERT trains its unsupervised language model using a Masked Language Model and Next Sentence Prediction. A common problem in NLP is the lack of enough training data. BERT can be pre-trained to learn general or specific language models using very large amounts of unlabelled text (e.g. web content, Wikipedia, etc.), and this knowledge can be transferred to a different downstream task in a process that receives the name \textit{fine-tuning}.

\newcite{devlin2018bert} have used fine-tuning to achieve state-of-the-art results on a wide variety of challenging natural language tasks, such as text classification, Question Answering (QA) and Named Entity Recognition and Classification (NERC). BERT has also been used successfully by other community practitioners for a wide range of NLP-related tasks \cite[among others]{liu2019fine,nogueira2019passage}.

Regarding the task of data anonymisation in particular, anonymisation systems may follow different approaches and pursue different objectives (Cormode and Srivastava, 2009). The first objective of these systems is to detect and classify the sensitive information contained in the documents to be anonymised. In order to achieve that, they use rule-based approaches, Machine Learning (ML) approaches, or a combination of both.

Although most of these efforts are for English texts --see, among others, the i2b2 de-identification challenges \cite{uzuner2007evaluating,stubbs2015automated}, \newcite{dernon2016deep}, or \newcite{khin2018deep}--, other languages are also attracting growing interest. Some examples are \newcite{mamede2016automated} for Portuguese and \newcite{tveit2004anonymization} for Norwegian. With respect to the anonymisation of text written in Spanish, recent studies include \newcite{medina2018building}, \newcite{hassan2018anonimizacion} and \newcite{garcia2018automating}. Most notably, in 2019 the first community challenge about anonymisation of medical documents in Spanish, MEDDOCAN\footnote{\href{http://temu.bsc.es/meddocan/}{http://temu.bsc.es/meddocan/}} \cite{marimon2019automatic}, was held as part of the IberLEF initiative. The winners of the challenge --the Neither-Language-nor-Domain-Experts (NLNDE) \cite{lange2019nlnde}-- achieved F1-scores as high as 0.975 in the task of sensitive information detection and categorisation by using recurrent neural networks with Conditional Random Field (CRF) output layers.

At the same challenge, \newcite{mao2019hadoken} occupied the 8\textsuperscript{th} position among 18 participants using BERT. According to the description of the system, the authors used BERT-Base Multilingual Cased and an output CRF layer. However, their system is $\sim$3 F1-score points below our implementation without the CRF layer.

%
%
\section{Materials and Methods}
\label{sec:materials-and-methods}

The aim of this paper is to evaluate BERT's multilingual model and compare it to other established machine-learning algorithms in a specific task: sensitive data detection and classification in Spanish clinical free text. This section describes the data involved in the experiments and the systems evaluated. Finally, we introduce the experimental setup.

%
%
\subsection{Data}
\label{ssec:data}

Two datasets are exploited in this article. Both datasets consist of plain text containing clinical narrative written in Spanish, and their respective manual annotations of sensitive information in BRAT \cite{stenetorp2012brat} standoff format\footnote{\href{https://brat.nlplab.org/standoff.html}{https://brat.nlplab.org/standoff.html}}. In order to feed the data to the different algorithms presented in Section \ref{ssec:systems}, these datasets were transformed to comply with the commonly used BIO sequence representation scheme \cite{Ramshaw1999}.

%
%
\subsubsection{\snubes}
\label{sssec:data-nubes}

\textsc{NUBes} \cite{lima2019a} is a corpus of around 7,000 real medical reports written in Spanish and annotated with negation and uncertainty information. Before being published, sensitive information had to be manually annotated and replaced for the corpus to be safely shared. In this article, we work with the \textsc{NUBes} version prior to its anonymisation, that is, with the manual annotations of sensitive information. It follows that the version we work with is not publicly available and, due to contractual restrictions, we cannot reveal the provenance of the data. In order to avoid confusion between the two corpus versions, we henceforth refer to the version relevant in this paper as \snubes\ (from `\textsc{NUBes} with Personal Health Information').

\snubes\ consists of 32,055 sentences annotated for 11 different sensitive information categories. Overall, it contains 7,818 annotations. The corpus has been randomly split into train (72\%), development (8\%) and test (20\%) sets to conduct the experiments described in this paper. The size of each split and the distribution of the annotations can be consulted in Tables \ref{tab:nubes-size} and \ref{tab:nubes-distribution}, respectively.

\begin{table}[!htbp]
    \centering
    \begin{tabular}{lrrr}
    \toprule
           & \multicolumn{1}{c}{train} & \multicolumn{1}{c}{dev} & \multicolumn{1}{c}{test} \\
    \midrule
    \# sentences & 23,079 & 2,565 & 6,411 \\
    \# tokens & 379,401 & 41,936 & 107,024 \\
    vocabulary  & 25,304 & 7,483 & 12,750 \\
    \# annotations & 5,562 & 677 & 1,579 \\
    \bottomrule
    \end{tabular}
    \caption{Size of the \snubes\ corpus}
    \label{tab:nubes-size}
\end{table}

The majority of sensitive information in \snubes\ are temporal expressions (`Date' and `Time'), followed by healthcare facility mentions (`Hospital'), and the age of the patient. Mentions of people are not that frequent, with physician names (`Doctor') occurring much more often than patient names (`Patient'). The least frequent sensitive information types, which account for $\sim$10\% of the remaining annotations, consist of the patient's sex, job, and kinship, and locations other than healthcare facilities (`Location'). Finally, the tag `Other' includes, for instance, mentions to institutions unrelated to healthcare and whether the patient is right- or left-handed. It occurs just 36 times.

\begin{table}[!htbp]
    \centering
    \begin{tabular}{lrrrrrr}
        \toprule
        & \multicolumn{2}{c}{train} & \multicolumn{2}{c}{dev} & \multicolumn{2}{c}{test} \\
        \midrule
        & \multicolumn{1}{c}{\#} & \multicolumn{1}{c}{\%} & \multicolumn{1}{c}{\#} & \multicolumn{1}{c}{\%} & \multicolumn{1}{c}{\#} & \multicolumn{1}{c}{\%} \\
        \midrule
        Date     & 2,165 & 39  & 251 & 37  & 660   & 41  \\
        Hospital & 1,012 & 18  & 105 & 16  & 275   & 17  \\
        Age      & 701   & 13  & 133 & 20  & 200   & 13  \\
        Time     & 608   & 11  & 63  & 9   & 155   & 10  \\
        Doctor   & 486   & 9   & 44  & 6   & 134   & 8   \\
        Sex      & 270   & 5   & 35  & 5   & 71    & 4   \\
        Kinship  & 158   & 3   & 20  & 3   & 44    & 3   \\
        Location & 71    & 1   & 10  & 1   & 19    & 1   \\
        Patient  & 48    & 1   & 5   & 1   & 11    & 1   \\
        Job      & 31    & 1   & 3   & 0   & 9     & 1   \\
        Other    & 12    & 0   & 8   & 1   & 16    & 1   \\
        \midrule
        Total    & 5,562 & 100 & 677 & 100 & 1,579 & 100 \\
        \bottomrule
    \end{tabular}
    \caption{Label distribution in the \snubes\ corpus}
    \label{tab:nubes-distribution}
\end{table}

%
%
\subsubsection{The MEDDOCAN corpus}
\label{sssec:data-meddocan}

The organisers of the MEDDOCAN shared task \cite{marimon2019automatic} curated a synthetic corpus of clinical cases enriched with sensitive information by health documentalists. In this regard, the MEDDOCAN evaluation scenario could be said to be somewhat far from the real use case the technology developed for the shared task is supposed to be applied in. However, at the moment it also provides the only public means for a rigorous comparison between systems for sensitive health information detection in Spanish texts.

The size of the MEDDOCAN corpus is shown in Table \ref{tab:meddocan-size}. Compared to \snubes\ (Table \ref{tab:nubes-size}), this corpus contains more sensitive information annotations, both in absolute and relative terms.

\begin{table}[!htbp]
    \centering
    \begin{tabular}{lrrr}
    \toprule
           & \multicolumn{1}{c}{train} & \multicolumn{1}{c}{dev} & \multicolumn{1}{c}{test} \\
    \midrule
    \# documents & 500 & 250 & 250 \\
    \# tokens & 360,407 & 138,812 & 132,961 \\
    vocabulary  & 26,355 & 15,985 & 15,397 \\
    \# annotations    & 11,333 & 5,801 & 5,661 \\
    \bottomrule
    \end{tabular}
    \caption{Size of the MEDDOCAN corpus}
    \label{tab:meddocan-size}
\end{table}

The sensitive annotation categories considered in MEDDOCAN differ in part from those in \snubes. Most notably, it contains finer-grained labels for location-related mentions --namely, `Address', `Territory', and  `Country'--, and other sensitive information categories that we did not encounter in \snubes\ (e.g., identifiers, phone numbers, e-mail addresses, etc.). In total, the MEDDOCAN corpus has 21 sensitive information categories. We refer the reader to the organisers' article \cite{marimon2019automatic} for more detailed information about this corpus. 

%
%
\subsection{Systems}
\label{ssec:systems}

Apart from experimenting with a pre-trained BERT model, we have run experiments with other systems and baselines, to compare them and obtain a better perspective about BERT's performance in these datasets. 

%
%
\subsubsection{Baseline}
\label{sssec:baseline}

As the simplest baseline, a sensitive data recogniser and classifier has been developed that consists of regular-expressions and dictionary look-ups. For each category to detect a specific method has been implemented. For instance, the Date, Age, Time and Doctor detectors are based on regular-expressions; Hospital, Sex, Kinship, Location, Patient and Job are looked up in dictionaries. The dictionaries are hand-crafted from the training data available, except for the Patient's case, for which the possible candidates considered are the 100 most common female and male names in Spain according to the Instituto Nacional de Estadística (INE; \textit{Spanish Statistical Office}).  

%
%
\subsubsection{CRF}
\label{sSsec:crf}

Conditional Random Fields (CRF) \cite{lafferty2001conditional} have been extensively used for tasks of sequential nature. 
In this paper, we propose as one of the competitive baselines a CRF classifier trained with sklearn-crfsuite\footnote{\href{https://sklearn-crfsuite.readthedocs.io}{https://sklearn-crfsuite.readthedocs.io}} for Python 3.5 and the following configuration: algorithm = \texttt{lbfgs}; maximum iterations = \texttt{100}; c1 = c2 = \texttt{0.1}; all transitions = \texttt{true}; optimise = \texttt{false}. The features extracted from each token are as follows:

\begin{itemize}[noitemsep]
    \item prefixes and suffixes of 2 and 3 characters;
    \item the length of the token in characters and the length of the sentence in tokens;
    \item whether the token is all-letters, a number, or a sequence of punctuation marks;
    \item whether the token contains the character `@';
    \item whether the token is the start or end of the sentence;
    \item the token's casing and the ratio of uppercase characters, digits, and punctuation marks to its length;
    \item and, the lemma, part-of-speech tag, and named-entity tag given by ixa-pipes\footnote{\href{https://ixa2.si.ehu.es/ixa-pipes}{https://ixa2.si.ehu.es/ixa-pipes}} \cite{agerri2014ixa} upon analysing the sentence the token belongs to.
\end{itemize}

Noticeably, none of the features used to train the CRF classifier is domain-dependent. However, the latter group of features is \textit{language} dependent.

%
%
\subsubsection{spaCy}
\label{sssec:spacy}

spaCy\footnote{\href{https://spacy.io}{https://spacy.io}} is a widely used NLP library that implements state-of-the-art text processing pipelines, including a sequence-labelling pipeline similar to the one described by \newcite{strubell2017fast}. spaCy offers several pre-trained models in Spanish, which perform basic NLP tasks such as Named Entity Recognition (NER). In this paper, we have trained a new NER model to detect \snubes\ labels. For this purpose, the new model uses all the labels of the training corpus coded with its context at sentence level. The network optimisation parameters and dropout values are the ones recommended in the documentation for small datasets\footnote{\href{https://spacy.io/usage/training}{https://spacy.io/usage/training}}. Finally, the model is trained using batches of size 64.
No more features are included, so the classifier is language-dependent but not domain-dependent. 

%
%
\subsubsection{BERT}
\label{sssec:bert}

\begin{figure}[!t]
  \includegraphics[width=\linewidth]{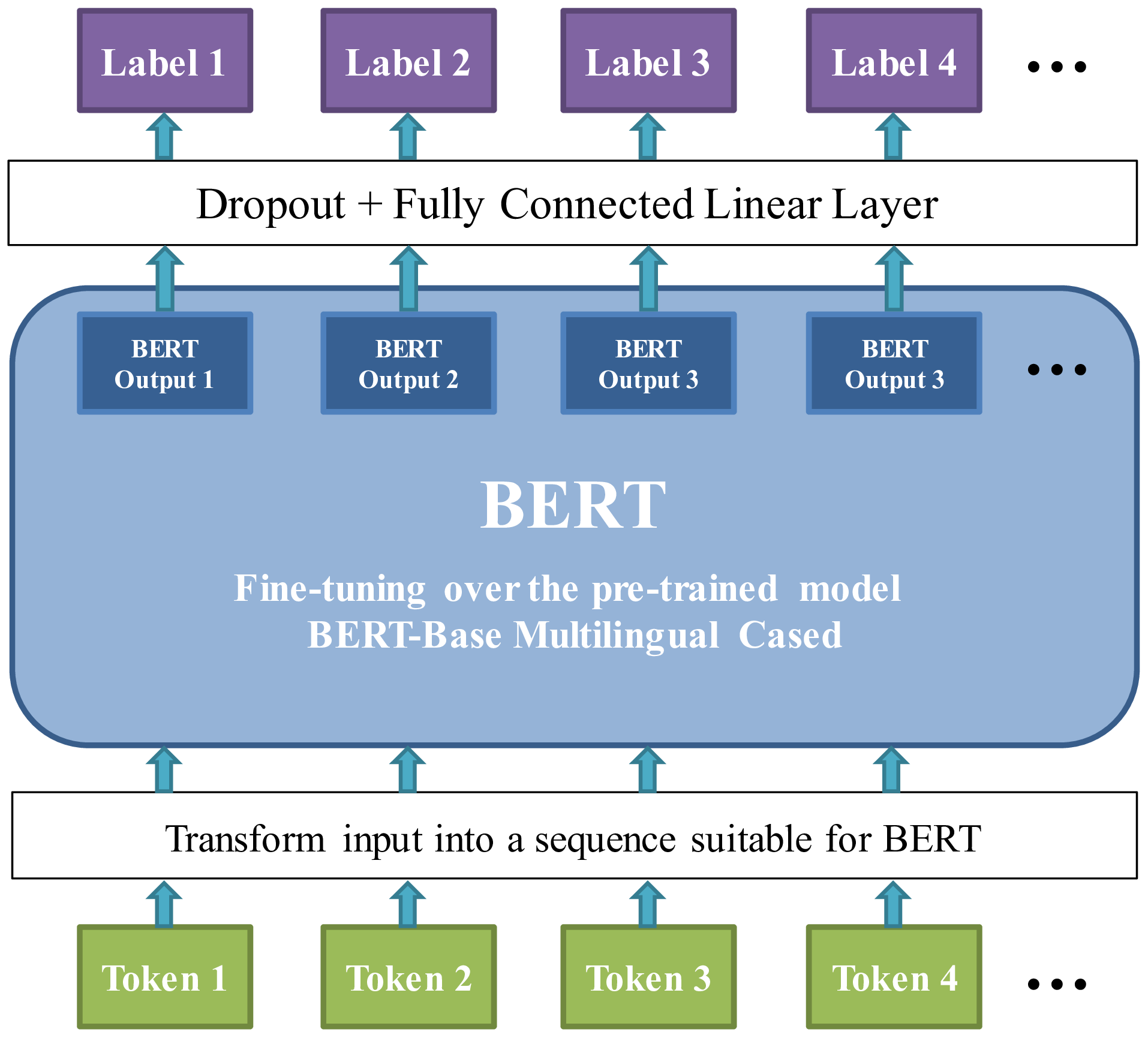}
  \caption{Pre-trained BERT with a Fully Connected layer on top to perform the fine-tuning}
  \label{fig:bert_diagram}
\end{figure}

As introduced earlier, BERT has shown an outstanding performance in NERC-like tasks, improving the start-of-the-art results for almost every dataset and language. We take the same approach here, by using the model BERT-Base Multilingual Cased\footnote{\href{https://github.com/google-research/bert/blob/master/multilingual.md}{https://github.com/google-research/bert}} with a Fully Connected (FC) layer on top to perform a fine-tuning of the whole model for an anonymisation task in Spanish clinical data. Our implementation is built on PyTorch\footnote{\href{https://pytorch.org}{https://pytorch.org}} and the PyTorch-Transformers library\footnote{\href{https://github.com/huggingface/transformers}{https://github.com/huggingface/transformers}} \cite{wolf2019transformers}. The training phase consists in the following steps (roughly depicted in Figure \ref{fig:bert_diagram}):

\begin{enumerate}
    \item \textbf{Pre-processing:} since we are relying on a pre-trained BERT model, we must match the same configuration by using a specific tokenisation and vocabulary. BERT also needs that the inputs contains special tokens to signal the beginning and the end of each sequence.
    \item \textbf{Fine-tuning:} the pre-processed sequence is fed into the model. BERT outputs the contextual embeddings that encode each of the inputted tokens. This embedding representation for each token is fed into the FC linear layer after a dropout layer (with a 0.1 dropout probability), which in turn outputs the logits for each possible class. The cross-entropy loss function is calculated comparing the logits and the gold labels, and the error is back-propagated to adjust the model parameters.
\end{enumerate}

We have trained the model using an AdamW optimiser \cite{loshchilov2017decoupled} with the learning rate set to 3e-5, as recommended by \newcite{devlin2018bert}, and with a gradient clipping of 1.0. We also applied a learning-rate scheduler that warms up the learning rate from zero to its maximum value as the training progresses, which is also a common practice. For each experiment set proposed below, the training was run with an early-stopping patience of 15 epochs. Then, the model that performed best against the development set was used to produce the reported results. 

The experiments were run on a 64-core server with operating system Ubuntu 16.04, 250GB of RAM memory, and 4 \textit{GeForce RTX 2080} GPUs with 11GB of memory. The maximum sequence length was set at 500 and the batch size at 12. In this setting, each epoch --a full pass through all the training data-- required about 10 minutes to complete.

%
%
\subsection{Experimental design}
\label{ssec:design}

We have conducted experiments with BERT in the two datasets of Spanish clinical narrative presented in Section \ref{ssec:data} The first experiment set uses \snubes, a corpus of real medical reports manually annotated with sensitive information. Because this corpus is not publicly available, and in order to compare the BERT-based model to other related published systems, the second set of experiments uses the MEDDOCAN 2019 shared task competition dataset. The following sections provide greater detail about the two experimental setups.

\begin{table*}[!htbp]
  \centering
    \begin{tabular}{lrrrrrrrrrrrr}
    \toprule
          && \multicolumn{3}{c}{Detection} && \multicolumn{3}{c}{Classification (relaxed)} && \multicolumn{3}{c}{Classification (strict)} \\
    \midrule
          && \multicolumn{1}{c}{Prec} & \multicolumn{1}{c}{Rec} & \multicolumn{1}{c}{F1} && \multicolumn{1}{c}{Prec} & \multicolumn{1}{c}{Rec} & \multicolumn{1}{c}{F1} && \multicolumn{1}{c}{Prec} & \multicolumn{1}{c}{Rec} & \multicolumn{1}{c}{F1}\\
    \midrule
    baseline && 0.853          & 0.469          & 0.605          && 0.779 & 0.429 & 0.553 && 0.721 & 0.396 & 0.512 \\
    CRF      && \textbf{0.968} & 0.921          & 0.944          && \textbf{0.952} & 0.907 & 0.929 && \textbf{0.941} & 0.896 & 0.918 \\
    spaCy    && 0.964          & 0.938          & 0.951          && 0.942 & 0.852 & 0.895 && 0.942 & 0.852 & 0.895 \\
    BERT     && 0.952          & \textbf{0.979} & \textbf{0.965} && 0.938 & \textbf{0.963} & \textbf{0.950} && 0.925 & \textbf{0.950} & \textbf{0.937} \\
    \bottomrule
    \end{tabular}
    \caption{Results of Experiment A: \snubes}
    \label{tab:nubes_scores}
\end{table*}

%
%
\subsubsection{Experiment A: \snubes}
\label{sssec:exp-nubes}

In this experiment set, we evaluate all the systems presented in Section \ref{ssec:systems}, namely, the rule-based baseline, the CRF classifier, the spaCy entity tagger, and BERT. The evaluation comprises three scenarios of increasing difficulty:

\begin{description}[noitemsep]

\item[Detection]- Evaluates the performance of the systems at predicting whether each token is sensitive or non-sensitive; that is, the measurements only take into account whether a sensitive token has been recognised or not, regardless of the BIO label and the category assigned. This scenario shows how good a system would be at obfuscating sensitive data (e.g., by replacing sensitive tokens with asterisks).

\item[Classification (relaxed)]- We measure the performance of the systems at predicting the sensitive information type of each token --i.e., the 11 categories presented in Section \ref{sssec:data-nubes} or `out'. Detecting entity types correctly is important if a system is going to be used to replace sensitive data by fake data of the same type (e.g., random people names).

\item[Classification (strict)]- This is the strictest evaluation, as it takes into account both the BIO label and the category assigned to each individual token. Being able to discern between two contiguous sensitive entities of the same type is relevant not only because it is helpful when producing fake replacements, but because it also yields more accurate statistics of the sensitive information present in a given document collection.

\end{description}

The systems are evaluated in terms of micro-average precision, recall and F1-score in all the scenarios.

In addition to the scenarios proposed, a subject worth being studied is the need of labelled data. Manually labelled data is an scarce and expensive resource, which for some application domains or languages is difficult to come by. In order to obtain an estimation of the dependency of each system on the available amount of training data, we have retrained all the compared models using decreasing amounts of data --from 100\% of the available training instances to just 1\%. The same data subsets have been used to train all the systems. Due to the knowledge transferred from the pre-trained BERT model, the BERT-based model is expected to be more robust to data scarcity than those that start their training from scratch.

%
%
\subsubsection{Experiment B: MEDDOCAN}
\label{ssec:exp-meddocan}

In this experiment set, our BERT implementation is compared to several systems that participated in the MEDDOCAN challenge: a CRF classifier \cite{perez2019vicomtech}, a spaCy entity recogniser \cite{perez2019vicomtech}, and NLNDE \cite{lange2019nlnde}, the winner of the shared task and current state of the art for sensitive information detection and classification in Spanish clinical text. Specifically, we include the results of a domain-independent NLNDE model (S2), and the results of a model enriched with domain-specific embeddings (S3). Finally, we include the results obtained by \newcite{mao2019hadoken} with a CRF output layer on top of BERT embeddings. MEDDOCAN consists of two scenarios:

\begin{description}[noitemsep]

\item[Detection]- This evaluation measures how good a system is at detecting sensitive text spans, regardless of the category assigned to them.  

\item[Classification]- In this scenario, systems are required to match exactly not only the boundaries of each sensitive span, but also the category assigned. 

\end{description}

The systems are evaluated in terms of micro-averaged precision, recall and F-1 score. Note that, in contrast to the evaluation in Experiment A, MEDDOCAN measurements are entity-based instead of tokenwise. An exhaustive explanation of the MEDDOCAN evaluation procedure is available online\footnote{\href{http://temu.bsc.es/meddocan/index.php/evaluation/}{http://temu.bsc.es/meddocan/index.php/evaluation/}}, as well as
the official evaluation script\footnote{\href{https://github.com/PlanTL-SANIDAD/MEDDOCAN-Evaluation-Script}{https://github.com/PlanTL-SANIDAD/MEDDOCAN-Evaluation-Script}}, which we used to obtain the reported results.

%
%
\section{Results}
\label{sec:results}

This section describes the results obtained in the two sets of experiments: \snubes\ and MEDDOCAN.

%
%
\subsection{Experiment A: \snubes}
\label{ssec:res-nubes}

Table \ref{tab:nubes_scores} shows the results of the conducted experiments in \snubes\ for all the compared systems. The included baseline serves to give a quick insight about how challenging the data is. With simple regular expressions and gazetteers a precision of 0.853 is obtained. On the other hand, the recall, which directly depends on the coverage provided by the rules and resources, drops to 0.469. Hence, this task is unlikely to be solved without the generalisation capabilities provided by machine-learning and deep-learning models.

Regarding the detection scenario --that is, the scenario concerned with a binary classification to determine whether each individual token conveys sensitive information or not--, it can be observed that BERT outperforms its competitors. A fact worth highlighting is that, according to these results, BERT achieves a precision lower than the rest of the systems (i.e., it makes more false positive predictions); in exchange, it obtains a remarkably higher recall. Noticeably, it reaches a recall of 0.979, improving by more than 4 points the second-best system, spaCy.

The table also shows the results for the relaxed metric that only takes into account the entity type detected, regardless of the BIO label (i.e., ignoring whether the token is at the beginning or in the middle of a sensitive sequence of tokens). The conclusions are very similar to those extracted previously, with BERT gaining 2.1 points of F1-score over the CRF based approach. 
\begin{table}[!b]
        
        \setlength{\fboxsep}{1.1mm} 
        \setlength{\tabcolsep}{0pt}
        \scriptsize
        \centering
        
        \begin{subtable}{\linewidth}
        \begin{tabular}{cr*{12}{G}}
            &          & \multicolumn{12}{r}{\scriptsize predicted} \smallskip \\
            &          & \rotz{Dat} & \rotz{Hos} & \rotz{Age} & \rotz{Tim} & \rotz{Doc} & \rotz{Sex} & \rotz{Kin} & \rotz{Loc} & \rotz{Pat} & \rotz{Job} & \rotz{Oth} &  \rotz{O} \\ 
            \multirow{12}{0.25cm}[-2.75cm]{\rotatebox{90}{\scriptsize true}}
            & \fc{Dat} & 0.90 & 0.00 & 0.01 & 0.01 & 0.00 & 0.00 & 0.00 & 0.00 & 0.00 & 0.00 & 0.00 & 0.08 \\
            & \fc{Hos} & 0.00 & 0.89 & 0.00 & 0.00 & 0.00 & 0.00 & 0.00 & 0.01 & 0.00 & 0.00 & 0.00 & 0.11 \\
            & \fc{Age} & 0.00 & 0.00 & 0.96 & 0.00 & 0.00 & 0.00 & 0.00 & 0.00 & 0.00 & 0.00 & 0.00 & 0.03 \\
            & \fc{Tim} & 0.01 & 0.00 & 0.00 & 0.95 & 0.00 & 0.00 & 0.00 & 0.00 & 0.00 & 0.00 & 0.00 & 0.04 \\
            & \fc{Doc} & 0.00 & 0.01 & 0.00 & 0.00 & 0.93 & 0.00 & 0.00 & 0.00 & 0.00 & 0.00 & 0.00 & 0.05 \\
            & \fc{Sex} & 0.00 & 0.00 & 0.00 & 0.00 & 0.00 & 1.00 & 0.00 & 0.00 & 0.00 & 0.00 & 0.00 & 0.00 \\
            & \fc{Kin} & 0.00 & 0.00 & 0.00 & 0.00 & 0.00 & 0.00 & 0.84 & 0.00 & 0.00 & 0.00 & 0.00 & 0.16 \\
            & \fc{Loc} & 0.00 & 0.08 & 0.00 & 0.00 & 0.00 & 0.00 & 0.00 & 0.35 & 0.00 & 0.08 & 0.00 & 0.58 \\
            & \fc{Pat} & 0.00 & 0.00 & 0.00 & 0.00 & 0.00 & 0.00 & 0.00 & 0.00 & 0.21 & 0.00 & 0.00 & 0.79 \\
            & \fc{Job} & 0.00 & 0.00 & 0.00 & 0.00 & 0.00 & 0.00 & 0.00 & 0.00 & 0.00 & 0.12 & 0.00 & 0.88 \\
            & \fc{Oth} & 0.00 & 0.00 & 0.00 & 0.00 & 0.00 & 0.00 & 0.00 & 0.00 & 0.00 & 0.00 & 0.00 & 1.00 \\
            & \fc{O}   & 0.00 & 0.00 & 0.00 & 0.00 & 0.00 & 0.00 & 0.00 & 0.00 & 0.00 & 0.00 & 0.00 & 1.00 \\
        \end{tabular}
        \caption{CRF}
        \label{tab:cm-crf}
        
    \end{subtable}
    
    \vspace{0.75\baselineskip}
    
    \begin{subtable}{\linewidth}
        \begin{tabular}{cr*{12}{G}}
            &          & \multicolumn{12}{r}{\scriptsize predicted} \smallskip \\
            &          & \rotz{Dat} & \rotz{Hos} & \rotz{Age} & \rotz{Tim} & \rotz{Doc} & \rotz{Sex} & \rotz{Kin} & \rotz{Loc} & \rotz{Pat} & \rotz{Job} & \rotz{Oth} &  \rotz{O} \\ 
            \multirow{12}{0.25cm}[-2.75cm]{\rotatebox{90}{\scriptsize true}}
            & \fc{Dat} &  0.93 & 0.00 & 0.01 & 0.01 & 0.00 & 0.00 & 0.00 & 0.00 & 0.00 & 0.00 & 0.00 & 0.05 \\
            & \fc{Hos} &  0.00 & 0.88 & 0.00 & 0.00 & 0.01 & 0.00 & 0.00 & 0.01 & 0.00 & 0.00 & 0.00 & 0.10 \\
            & \fc{Age} &  0.00 & 0.00 & 0.98 & 0.00 & 0.00 & 0.00 & 0.00 & 0.00 & 0.00 & 0.00 & 0.00 & 0.02 \\
            & \fc{Tim} &  0.01 & 0.00 & 0.00 & 0.95 & 0.00 & 0.00 & 0.00 & 0.00 & 0.00 & 0.00 & 0.00 & 0.04 \\
            & \fc{Doc} &  0.00 & 0.00 & 0.00 & 0.00 & 0.95 & 0.00 & 0.00 & 0.00 & 0.00 & 0.00 & 0.00 & 0.05 \\
            & \fc{Sex} &  0.00 & 0.00 & 0.00 & 0.00 & 0.00 & 1.00 & 0.00 & 0.00 & 0.00 & 0.00 & 0.00 & 0.00 \\
            & \fc{Kin} &  0.00 & 0.00 & 0.00 & 0.00 & 0.00 & 0.00 & 0.95 & 0.00 & 0.00 & 0.00 & 0.00 & 0.05 \\
            & \fc{Loc} &  0.00 & 0.15 & 0.00 & 0.00 & 0.00 & 0.00 & 0.00 & 0.27 & 0.00 & 0.08 & 0.00 & 0.58 \\
            & \fc{Pat} &  0.00 & 0.00 & 0.00 & 0.00 & 0.07 & 0.00 & 0.00 & 0.07 & 0.21 & 0.00 & 0.00 & 0.64 \\
            & \fc{Job} &  0.00 & 0.00 & 0.00 & 0.00 & 0.00 & 0.00 & 0.00 & 0.00 & 0.00 & 0.12 & 0.00 & 0.88 \\
            & \fc{Oth} &  0.00 & 0.00 & 0.00 & 0.00 & 0.00 & 0.00 & 0.00 & 0.00 & 0.00 & 0.00 & 0.00 & 1.00 \\
            & \fc{O} &  0.00 & 0.00 & 0.00 & 0.00 & 0.00 & 0.00 & 0.00 & 0.00 & 0.00 & 0.00 & 0.00 & 1.00 \\

        \end{tabular}
        \caption{spaCy}
        \label{tab:cm-spacy}
    \end{subtable}
    
    \vspace{0.75\baselineskip}
    
    \begin{subtable}{\linewidth}
        \begin{tabular}{cr*{12}{G}}
            &          & \multicolumn{12}{r}{\scriptsize predicted} \smallskip \\
            &          & \rotz{Dat} & \rotz{Hos} & \rotz{Age} & \rotz{Tim} & \rotz{Doc} & \rotz{Sex} & \rotz{Kin} & \rotz{Loc} & \rotz{Pat} & \rotz{Job} & \rotz{Oth} &  \rotz{O} \\ 
            \multirow{12}{0.25cm}[-2.75cm]{\rotatebox{90}{\scriptsize true}}
            & \fc{Dat} &  0.96 & 0.00 & 0.01 & 0.01 & 0.00 & 0.00 & 0.00 & 0.00 & 0.00 & 0.00 & 0.00 & 0.02 \\
            & \fc{Hos} &  0.00 & 0.96 & 0.00 & 0.00 & 0.01 & 0.00 & 0.00 & 0.01 & 0.00 & 0.00 & 0.00 & 0.03 \\
            & \fc{Age} &  0.00 & 0.00 & 0.99 & 0.00 & 0.00 & 0.00 & 0.00 & 0.00 & 0.00 & 0.00 & 0.00 & 0.00 \\
            & \fc{Tim} &  0.01 & 0.00 & 0.00 & 0.99 & 0.00 & 0.00 & 0.00 & 0.00 & 0.00 & 0.00 & 0.00 & 0.00 \\
            & \fc{Doc} &  0.00 & 0.00 & 0.00 & 0.00 & 1.00 & 0.00 & 0.00 & 0.00 & 0.00 & 0.00 & 0.00 & 0.00 \\
            & \fc{Sex} &  0.00 & 0.00 & 0.00 & 0.00 & 0.00 & 1.00 & 0.00 & 0.00 & 0.00 & 0.00 & 0.00 & 0.00 \\
            & \fc{Kin} &  0.00 & 0.00 & 0.00 & 0.00 & 0.00 & 0.00 & 1.00 & 0.00 & 0.00 & 0.00 & 0.00 & 0.00 \\
            & \fc{Loc} &  0.00 & 0.23 & 0.00 & 0.00 & 0.00 & 0.00 & 0.00 & 0.50 & 0.00 & 0.08 & 0.00 & 0.19 \\
            & \fc{Pat} &  0.00 & 0.00 & 0.00 & 0.00 & 0.00 & 0.00 & 0.00 & 0.00 & 1.00 & 0.00 & 0.00 & 0.00 \\
            & \fc{Job} &  0.00 & 0.00 & 0.00 & 0.00 & 0.00 & 0.00 & 0.00 & 0.00 & 0.00 & 0.41 & 0.00 & 0.59 \\
            & \fc{Oth} &  0.00 & 0.00 & 0.00 & 0.00 & 0.00 & 0.00 & 0.00 & 0.00 & 0.00 & 0.00 & 0.00 & 1.00 \\
            & \fc{O} &  0.00 & 0.00 & 0.00 & 0.00 & 0.00 & 0.00 & 0.00 & 0.00 & 0.00 & 0.00 & 0.00 & 1.00 \\
        \end{tabular}
        \caption{BERT}
        \label{tab:cm-bert}
    \end{subtable}
    
    \caption{Confusion matrices for the sensitive information classification task on the \snubes\ corpus}
    \label{tab:confusion-matrices}
    
\end{table}
The confusion matrices of the predictions made by CRF, spaCy, and BERT in this scenario are shown in Table \ref{tab:confusion-matrices}. As can bee seen, BERT has less difficulty in predicting correctly less frequent categories, such as `Location', `Job', and `Patient'. One of the most common mistakes according to the confusion matrices is classifying hospital names as `Location' instead of the more accurate `Hospital'; this is hardly a harmful error, given that a hospital is actually a location. Last, the category `Other' is completely leaked by all the compared systems, most likely due to its almost total lack of support in both training and evaluation datasets.

To finish with this experiment set, Table \ref{tab:nubes_scores} also shows the strict classification precision, recall and F1-score for the compared systems. Despite the fact that, in general, the systems obtain high values, BERT outperforms them again. BERT's F1-score is 1.9 points higher than the next most competitive result in the comparison. More remarkably, the recall obtained by BERT is about 5 points above.

Upon manual inspection of the errors committed by the BERT-based model, we discovered that it has a slight tendency towards producing ill-formed BIO sequences (e.g, starting a sensitive span with `Inside' instead of `Begin'; see Table \ref{tab:error-example}). We could expect that complementing the BERT-based model with a CRF layer on top would help enforce the emission of valid sequences, alleviating this kind of errors and further improving its results.

\begin{table}[!htbp]
    \centering
    \begin{subtable}{\linewidth}
    \begin{tabular}{rl}
        \toprule
        & \small \texttt{Acudirá a la Clínica Marseille \ } \\
        \midrule
        true & \small \texttt{\ \ \ O \ \ \ O B \ \ \ \ \textbf{I} \ \ \ \ \ \ \ I} \\
        predicted & \small \texttt{\ \ \ O \ \ \ O B \ \ \ \ \textbf{B} \ \ \ \ \ \ \ I} \\
        \bottomrule
    \end{tabular}
    \caption{Example 1: ``[The patient] will attend the Marseille Clinic''}
    \label{tab:error-example-1}
    \end{subtable}
    
    \vspace{0.75\baselineskip}
    
    \begin{subtable}{\linewidth}
    \begin{tabular}{rl}
        \toprule
        & \small \texttt{control ( 15 y 22 de junio ) \ \ \ } \\
        \midrule
        true & \small \texttt{\ \ \ O \ \ \ O B \ O \textbf{B} \ I \ \ \ I \ \ O} \\
        predicted & \small \texttt{\ \ \ O \ \ \ O B \ O \textbf{I} \ I \ \ \ I \ \  O} \\
        \bottomrule
    \end{tabular}
    \caption{Example 2: ``inspection (15 and 22 of june)''}
    \label{tab:error-example-2}
    \end{subtable}
    
    \vspace{0.75\baselineskip}
    
    \begin{subtable}{\linewidth}
    \begin{tabular}{rl}
        \toprule
        & \small \texttt{Niño de 4 años y medio \ \ \ \ \ \ \ \ \ } \\
        \midrule
        true & \small \texttt{\ \textbf{B} \ \ O \ B \ I \ \ \textbf{I} \ \ \textbf{I}} \\
        predicted & \small \texttt{\ \textbf{O} \ \ O \ B \ I \ \ \textbf{O} \ \ \textbf{O}} \\
        \bottomrule
    \end{tabular}
    \caption{Example 3: ``4 and a half years-old boy''}
    \label{tab:error-example-3}
    \end{subtable}
    \caption{BERT error examples (only BIO-tags are shown; differences between gold annotations and predictions are highlighted in bold)}
    \label{tab:error-example}
\end{table}

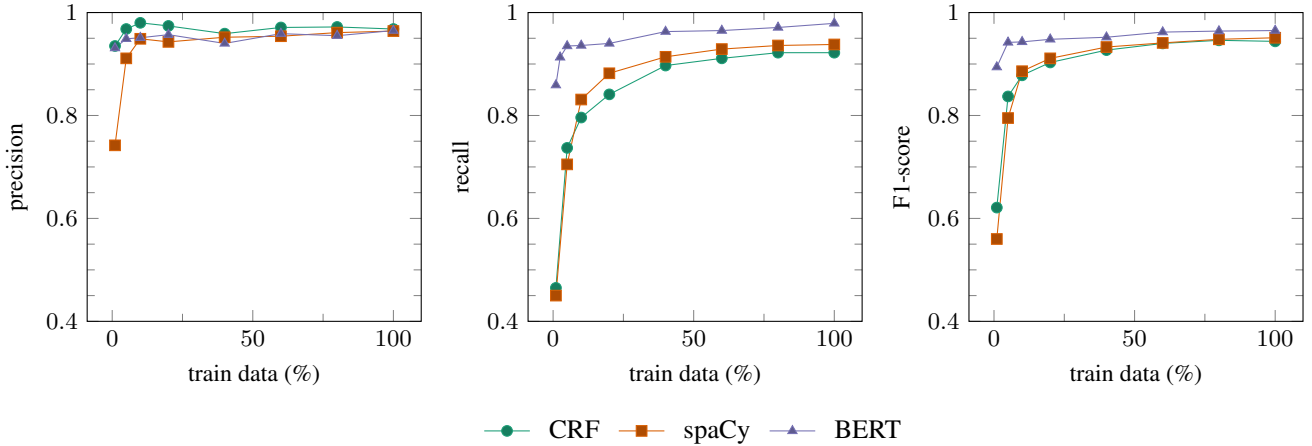
\begin{figure*}[!htbp]
    \centering

\begin{tikzpicture}

    \begin{groupplot}[
        group style = {group size = 3 by 1, horizontal sep = 40pt}, 
        width = 0.3475\linewidth, 
        height = 0.33\linewidth, 
        label style={font=\small},
        tick label style={font=\small},
        ymin=0.4,
        ymax=1.0,
        minor x tick num=1,
        minor y tick num=3,
        legend style={draw=none}
    ]
    
        \nextgroupplot[xlabel={train data (\%)},ylabel={precision},
        legend style = { column sep = 5pt, legend columns = -1, legend to name = grouplegend,}]

        \addplot coordinates {
            (1,0.935)
            (5,0.968)(10,0.980)(20,0.974)(40,0.959)(60,0.971)(80,0.972)(100,0.968)
        }; \addlegendentry{CRF}
        \addplot coordinates {
            (1,0.742)
            (5,0.911)(10,0.949)(20,0.943)(40,0.952)(60,0.954)(80,0.961)(100,0.964)
        }; \addlegendentry{spaCy}
        \addplot coordinates {
            (1, 0.931)
            (5,0.949)(10,0.951)(20,0.957)(40,0.940)(60,0.959)(80,0.955)(100,0.965)
        }; \addlegendentry{BERT}

        \nextgroupplot[xlabel={train data (\%)},ylabel={recall}]
        \addplot coordinates {
            (1,0.465)
            (5, 0.737)(10,0.796)(20,0.841)(40,0.897)(60,0.911)(80,0.922)(100,0.922)
        };
        \addplot coordinates {
            (1,0.450)
            (5,0.705)(10,0.831)(20,0.882)(40,0.914)(60,0.929)(80,0.936)(100,0.938)
        };
        \addplot coordinates {
            (1,0.859)(2.5,0.913)(5,0.935)(10,0.936)(20,0.940)(40,0.963)(60,0.965)(80,0.971)(100,0.979)
        };
        
        \nextgroupplot[xlabel={train data (\%)},ylabel={F1-score}]
        \addplot coordinates {
            (1,0.621)
            (5,0.837)(10,0.878)(20,0.903)(40,0.927)(60,0.940)(80,0.946)(100,0.944)
        };
        \addplot coordinates {
            (1,0.560)
            (5,0.795)(10,0.886)(20,0.911)(40,0.933)(60,0.941)(80,0.948)(100,0.951)
        };
        \addplot coordinates {
            (1,0.894)
            (5,0.942)(10,0.943)(20,0.948)(40,0.952)(60,0.962)(80,0.964)(100,0.965)
        };
    
    \end{groupplot}
    
    \node at ($(group c2r1) + (0,-3.5cm)$) {\ref{grouplegend}}; 
    
\end{tikzpicture}

    \caption{Performance decay with decreasing amounts of training data on the sensitive information detection task in the \snubes\ corpus}
    \label{fig:binary_nubes_data_drop}
\end{figure*}

\begin{table*}[!htbp]
    \centering
    \begin{tabular}{lrrrrrrrr}
    \toprule
                                            && \multicolumn{3}{c}{Detection} && \multicolumn{3}{c}{Classification} \\
    \midrule
                                            && \multicolumn{1}{c}{Prec} & \multicolumn{1}{c}{Rec} & \multicolumn{1}{c}{F1} && \multicolumn{1}{c}{Prec} & \multicolumn{1}{c}{Rec} & \multicolumn{1}{c}{F1}\\
    \midrule
    CRF \cite{perez2019vicomtech}           && \textbf{0.977} & 0.943 & 0.960 && \textbf{0.971} & 0.937 & 0.954 \\
    spaCy \cite{perez2019vicomtech}         && 0.967 & 0.953 & 0.965 && 0.965 & 0.947 & 0.956 \\
    NLND S2 \cite{lange2019nlnde}           && 0.976 & 0.973 & 0.974 && \textbf{0.971} & 0.968 & \textbf{0.970} \\
    NLND S3 \cite{lange2019nlnde}           && 0.975 & \textbf{0.975} & \textbf{0.975} && 0.970 & \textbf{0.969} & \textbf{0.970} \\
    BERT + CRF \cite{mao2019hadoken} && 0.968 & 0.919 & 0.943 && 0.965 & 0.912 & 0.937 \\
    BERT                         && 0.973 & 0.972 & 0.972 && 0.968 & 0.967 & 0.967 \\
    \bottomrule
    \end{tabular}
    \caption{Results of Experiment B: MEDDOCAN}
    \label{tab:meddocan_scores}
\end{table*}

Finally, Figure \ref{fig:binary_nubes_data_drop} shows the impact of decreasing the amount of training data in the detection scenario. It shows the difference in precision, recall, and F1-score with respect to that obtained using 100\% of the training data. A general downward trend can be observed, as one would expect: less training data leads to less accurate predictions. However, the BERT-based model is the most robust to training-data reduction, showing an steadily low performance loss. With 1\% of the dataset (230 training instances), the BERT-based model only suffers a striking 7-point F1-score loss, in contrast to the 32 and 39 points lost by the CRF and spaCy models, respectively. This steep performance drop stems to a larger extent from recall decline, which is not that marked in the case of BERT. Overall, these results indicate that the transfer-learning achieved through the BERT multilingual pre-trained model not only helps obtain better results, but also lowers the need of manually labelled data for this application domain.

%
%
\subsection{Experiment B: MEDDOCAN}
\label{ssec:res-meddocan}

The results of the two MEDDOCAN scenarios --detection and classification-- are shown in Table \ref{tab:meddocan_scores}. These results follow the same pattern as in the previous experiments, with the CRF classifier being the most precise of all, and BERT outperforming both the CRF and spaCy classifiers thanks to its greater recall. We also show the results of \newcite{mao2019hadoken} who, despite of having used a BERT-based system, achieve lower scores than our models. The reason why it should be so remain unclear.

With regard to the winner of the MEDDOCAN shared task, the BERT-based model has not improved the scores obtained by neither the domain-dependent (S3) nor the domain-independent (S2) NLNDE model. However, attending to the obtained results, BERT remains only 0.3 F1-score points behind, and would have achieved the second position among all the MEDDOCAN shared task competitors. Taking into account that only 3\% of the gold labels remain incorrectly annotated, the task can be considered almost solved, and it is not clear if the differences among the systems are actually significant, or whether they stem from minor variations in initialisation or a long-tail of minor labelling inconsistencies.

%
%
\section{Conclusions and Future Work}
\label{sec:conclusions}

In this work we have briefly introduced the problems related to data privacy protection in clinical domain.
We have also described some of the groundbreaking advances on the Natural Language Processing field due to the appearance of Transformers-based deep-learning architectures and transfer learning from very large general-domain multilingual corpora, focusing our attention in one of its most representative examples, Google's BERT model.

In order to assess the performance of BERT for Spanish clinical data anonymisation, we have conducted several experiments with a BERT-based sequence labelling approach using the pre-trained multilingual BERT model shared by Google as the starting point for the model training. We have compared this BERT-based sequence labelling against other methods and systems. One of the experiments uses the MEDDOCAN 2019 shared task dataset, while the other uses a novel Spanish clinical reports dataset called \snubes.

The results of the experiments show that, in \snubes, the BERT-based model outperforms the other systems without requiring any adaptation or domain-specific feature engineering, just by being trained on the provided labelled data. Interestingly, the BERT-based model obtains a remarkably higher recall than the other systems. High recall is a desirable outcome because, when anonymising sensible documents, the accidental leak of sensible data is likely to be more dangerous than the unintended over-obfuscation of non-sensitive text.

Further, we have conducted an additional experiment on this dataset by progressively reducing the training data for all the compared systems. The BERT-based model shows the highest robustness to training-data scarcity, loosing only 7 points of F1-score when trained on 230 instances instead of 21,371. These observation are in line with the results obtained by the NLP community using BERT for other tasks.

The experiments with the MEDDOCAN 2019 shared task dataset follow the same pattern. In this case, the BERT-based model falls 0.3 F1-score points behind the shared task winning system, but it would have achieved the second position in the competition with no further refinement.

Since we have used a pre-trained multilingual BERT model, the same approach is likely to work for other languages just by providing some labelled training data. Further, this is the simplest fine-tuning that can be performed based on BERT. More sophisticated fine-tuning layers could help improve the results. For example, it could be expected that a CRF layer helped enforce better BIO tagging sequence predictions. Precisely, \newcite{mao2019hadoken} participated in the MEDDOCAN competition using a BERT+CRF architecture, but their reported scores are about 3 points lower than our implementation. From the description of their work, it is unclear what the source of this score difference could be.

Further, at the time of writing this paper, new multilingual pre-trained models and Transformer architectures have become available. It would not come as a surprise that these new resources and systems --e.g., XLM-RoBERTa \cite{conneau2019unsupervised} or BETO \cite{wu2019beto}, a BERT model fully pre-trained on Spanish texts-- further advanced the state of the art in this task.


%
%
\section{Acknowledgements}

This work has been supported by Vicomtech and partially funded by the project DeepReading (RTI2018-096846-B-C21, MCIU/AEI/FEDER,UE).

%
%
\section{Bibliographical References}
\label{sec:ref}

\bibliographystyle{lrec}
\bibliography{lrec2020hitzalmed}

\end{document}